\documentclass[letterpaper]{article} 
\usepackage{aaai25}  
\usepackage{times}  
\usepackage{helvet}  
\usepackage{courier}  
\usepackage[hyphens]{url}  
\usepackage{graphicx} 
\usepackage{natbib}  
\usepackage{caption} 
\usepackage{algorithm}
\usepackage{algorithmic}

\usepackage{multirow}
\usepackage{booktabs}

\usepackage{newfloat}
\usepackage{listings}
\DeclareCaptionStyle{ruled}{labelfont=normalfont,labelsep=colon,strut=off} 
\floatstyle{ruled}
\newfloat{listing}{tb}{lst}{}
\floatname{listing}{Listing}
\title{Enhancing Non-English Capabilities of English-Centric Large Language Models Through Deep Supervision Fine-Tuning}

\author{
    Wenshuai Huo\textsuperscript{\rm 1,2}, 
    Xiaocheng Feng\textsuperscript{\rm 1,2} \thanks{Corresponding author} , 
    Yichong Huang\textsuperscript{\rm 1}, 
    Chengpeng Fu\textsuperscript{\rm 1,2},  Baohang Li\textsuperscript{\rm 1}, \\Yangfan Ye\textsuperscript{\rm 1}, 
    Zhirui Zhang\textsuperscript{\rm 3}, Dandan Tu\textsuperscript{\rm 3}, Duyu Tang\textsuperscript{\rm 3}, Yunfei Lu\textsuperscript{\rm 3}, Hui Wang\textsuperscript{\rm 2}, Bing Qin\textsuperscript{\rm 1,2 } \footnotemark[1]\\
}
\affiliations{
    \textsuperscript{\rm 1}Harbin Institution of Technology \\
    \textsuperscript{\rm 2}Pengcheng Laboratory\\
    \textsuperscript{\rm 3}Huawei Technologies Co., Ltd\\

    \{wshuo, xcfeng, ychuang, cpfu, baohangli, yfye, qinb\}@ir.hit.edu.cn, 
    wangh06@pcl.ac.cn\\
    \{tudandan, tangduyu, luyunfei6\}@huawei.com, zrustc11@gmail.com

}

\usepackage{bibentry}

\begin{document}

\maketitle

\begin{abstract}

Large language models (LLMs) have demonstrated significant progress in multilingual language understanding and generation. However, due to the imbalance in training data, their capabilities in non-English languages are limited.
Recent studies revealed the English-pivot multilingual mechanism of LLMs, 
where LLMs implicitly convert non-English queries into English ones at the bottom layers and adopt English for thinking at the middle layers. 
However, due to the absence of explicit supervision for cross-lingual alignment in the intermediate layers of LLMs, the internal representations during these stages may become inaccurate.
In this work, we introduce a deep supervision fine-tuning method (DFT) that incorporates additional supervision over the internal layers of the model to guide its workflow. 
Specifically, we introduce two training objectives on different layers of LLMs: one at the bottom layers to constrain the conversion of the target language into English, and another at the middle layers to constrain reasoning in English. To effectively achieve the guiding purpose, we designed two types of supervision signals: logits and feature, which represent a stricter constraint and a relatively more relaxed guidance.
Our method guides the model to not only consider the final generated result when processing non-English inputs but also ensure the accuracy of internal representations.
We conducted extensive experiments on typical English-centric LLMs, LLaMA-2 and Gemma-2.
The results on 8 multilingual datasets show that our method significantly outperforms traditional fine-tuning methods.

\end{abstract}

\begin{figure}[!t]
  \centering
    \includegraphics[clip,width=0.9\columnwidth,]{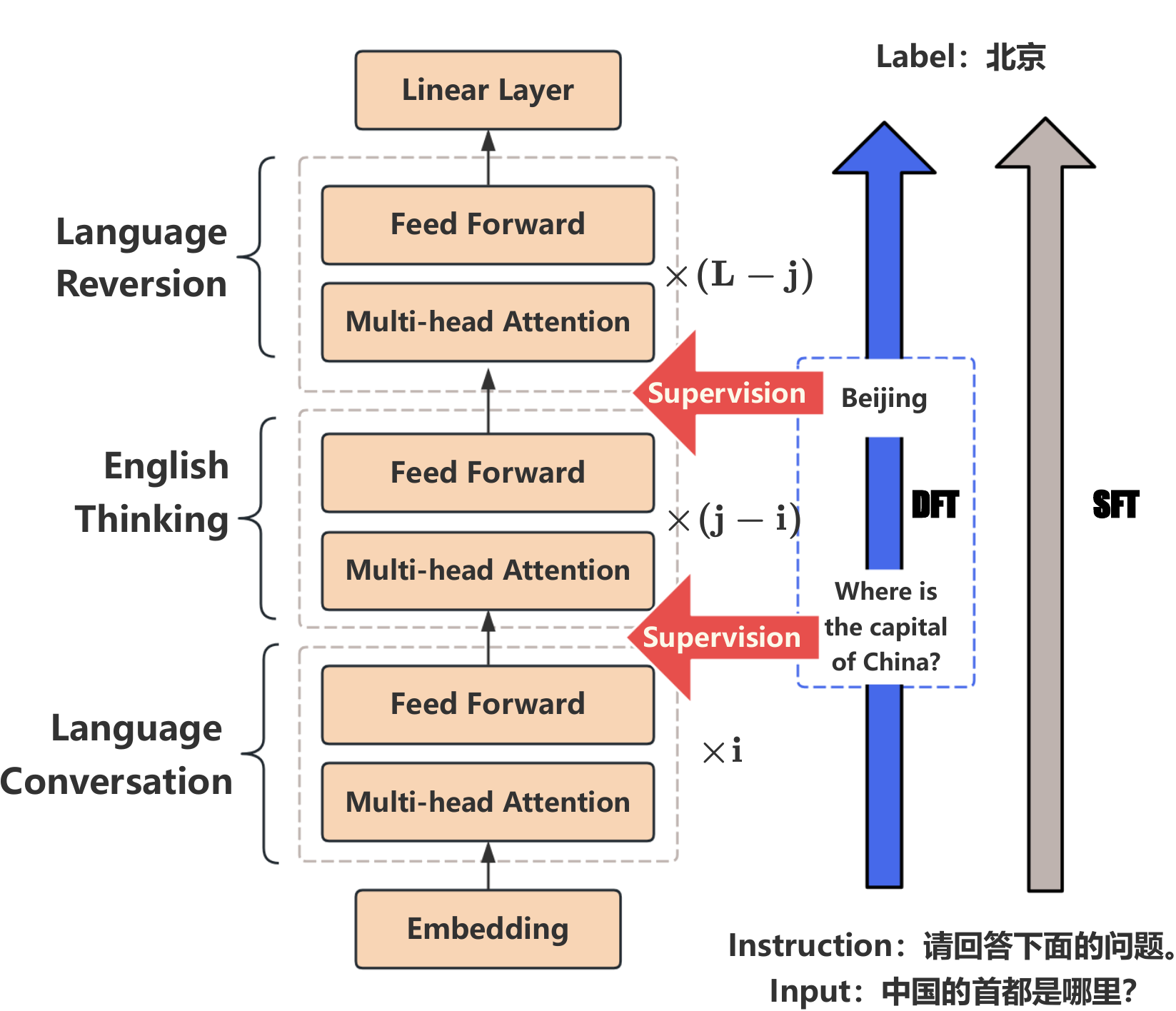}
    \caption{The illustration of Depth Supervision Fine-Tuning (DFT) and Baseline Methods. The left side represents an English-dominated large language model, which can be divided into three parts from shallow to deep layers: Language Conversion, English Thinking, and Language Reversion. The right side shows an sample of Chinese instruction tuning. Blue arrows represent the DFT method, while gray arrows represent the Baseline method. Traditional fine-tuning methods focus only on the model predicting the corresponding target output based on the input instruction. In contrast, our method adds supervision to the process, explicitly guiding the model's workflow when processing non-English inputs.
    }
  \label{fig:fig1}
\end{figure}

\section{Introduction}

The development of large language models has achieved revolutionary breakthroughs in the field of natural language processing, demonstrating exceptional performance \cite{openai2023chatgpt,touvron2023llama,team2023gemini}. However, performance disparities still exist across different languages due to the imbalance in the training data \cite{nguyen2023democratizing,huang2023not,zhu2023multilingual}. For instance, the well-known LLaMA series models are trained on over 90 \% English data. This results in a significant gap between their multilingual capabilities and their performance in English.

To enhance the non-English processing capabilities of these models, researchers have made numerous attempts. A common and effective approach is to fine-tune pre-trained models on instruction datasets in the target languages \cite{zhu2023extrapolating,li2023bactrian,zhang2024enhancing,li2024x,zhao2024llama}.
As research advances, some studies have explored the internal mechanisms behind how large models handle multilingualism, revealing an English-centric workflow when processing non-English inputs.
Specifically, \citet{zhao2024large} analyzed activation patterns in specific languages and inferred that the bottom layers of the model convert input from various languages into English, while the top layers perform the reverse conversion. \citet{wendler2024llamas} found that by decoding early in the model's intermediate layers, rather than the final layer, LLMs tend to use English as an internal pivot language when processing multilingual inputs.
These studies suggest that LLMs dominated by English can be divided into three stages when processing non-English inputs: converting non-English to English (or an English representation space), thinking in English, and converting English back to the target language. However, due to the absence of explicit supervision for cross-lingual alignment in the intermediate layers of LLMs, the internal representations during these stages may become inaccurate. To the best of our knowledge, no studies have yet proposed targeted optimization schemes for this phenomenon.

To tackle the aforementioned problem, we propose the deep supervision fine-tuning method (DFT) that aims to explicitly guide the LLMs' performance in the three stages when processing non-English inputs. Specifically, DFT incorporates additional supervision over the internal layers of the model.
As shown in Figure \ref{fig:fig1}, unlike traditional fine-tuning methods, DFT constrains not only the final output of the model but also the intermediate process. In the bottom layers of the model, DFT guides the conversion from non-English to English. In the middle layers, DFT guides the model to obtain answers in the English space. To effectively constrain the intermediate process, we propose two supervision schemes based on logits and features. Additionally, to accurately identify the critical layers at different stages, we propose an entropy-based selection strategy.

We conducted extensive experiments on 8 commonly used multilingual benchmarks and the results demonstrate the effectiveness of our method. Specifically, for multilingual QA tasks where both input and output are in the target language, our method achieved significant improvements.

\section{Background}

\subsubsection{Instruction Fine-Tuning.}
The technology of instruction tuning enables pre-trained LLMs to comprehend instructions and handle downstream tasks effectively via the training on an annotated instruction dataset $ D = \{(x_i, y_i) \}_{i=1}^{N} $, where $ x $ is the input question and $ y $ is the expected output answer. The training objective is to minimize the following negative log-likelihood:
\begin{equation}
    \label{eq:IFT}
    \mathcal{L}(\theta) = \sum_{i=1}^{N} - \log P(y_i|x_i;\theta),
\end{equation}
where $\theta$ denotes the learnable parameters of the model. However, most instruct datasets are in English, which limits the potential of large models to address tasks in non-English languages.

\subsubsection{Multilingual Instruction Fine-Tuning.}

To enhance the multilingual capabilities of large language models, a common approach is to translate English instruction data into the target language (tgt), creating a dataset $ D^{tgt} = \{(x_{i}^{tgt}, y_{i}^{tgt}) \}_{i=1}^{N} $. The model is then fine-tuned using the translated instruction data. Similar to equation \ref{eq:IFT}, the loss function $\mathcal{L}_{TFT}$ is:
\begin{equation}
    \label{eq:MT-IFT}
    \mathcal{L}_{TFT}(\theta) = \sum_{i=1}^{N} - \log P(y_{i}^{tgt}|x_{i}^{tgt};\theta).
\end{equation}

In this work, we also fine-tune the model using multilingual instruction datasets constructed through translation. However, unlike traditional fine-tuning methods, we focus not only on the model's final predictions but also impose constraints to ensure the model achieves accurate intermediate results.

\subsubsection{Deep Supervision Networks.}
Deep Supervision Networks (DSNs) represent a special training strategy that optimizes the training process by incorporating supervision at multiple intermediate layers of the model. Unlike traditional deep learning architectures that compute the loss only at the final layer $L$, DSNs also calculate loss at several hidden layers $k$. The total loss function $\mathcal{L}$ can be expressed as: 
\begin{equation}
    \label{eq:DSN}
    \mathcal{L}(\theta) = \mathcal{L}_L(\theta) + \alpha\mathcal{L}_k(\theta_{1:k}),
\end{equation}
$\theta_{1:k}$ denotes only  the parameters of layer $k$ to layer $1$ are learned and updated by minimising the intermediate loss ($\mathcal{L}_k$); $\alpha$ is a hyper-parameter to control the balance between the intermediate
supervision loss ($\mathcal{L}_k$) and the final output layer loss ($\mathcal{L}_L$).

In this work, we take inspiration from the training strategy of DSNs by introducing additional supervision over the intermediate layers of the model. Unlike traditional DSNs, where the intermediate supervision is consistent with the final output target \cite{li2022comprehensive}, our additional supervision is differ from the final output. The goal is to guide the workflow of LLMs when processing non-English inputs.

\begin{figure*}[t]
  \centering
    \includegraphics[clip,width=2.0\columnwidth,]{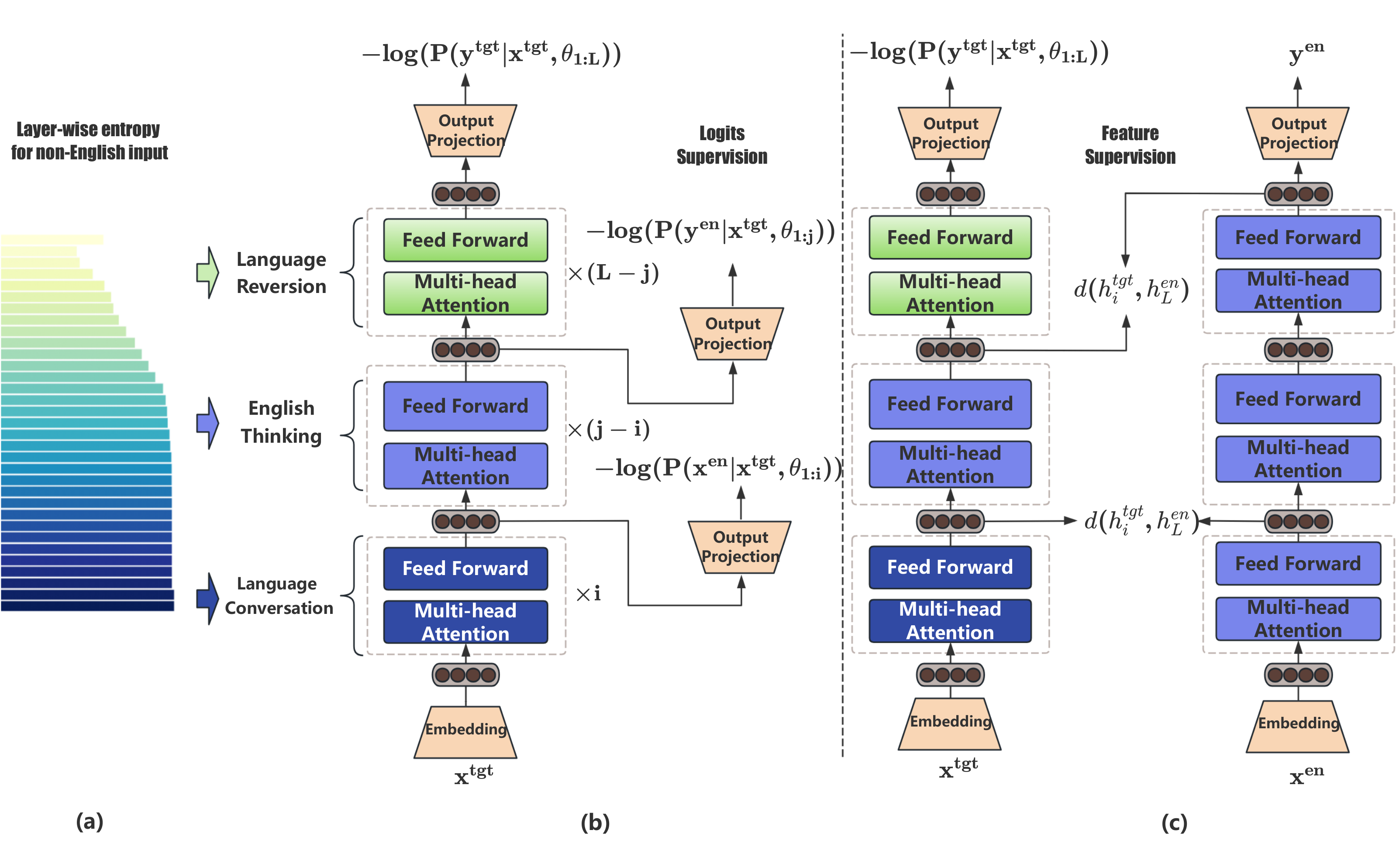}
    \caption{The illustration of the proposed methods DFT-logits (b) and DFT-feature (c). The heatmap (a) represents the entropy values of each layer in English-dominated Large Language Model when processing non-English inputs.  The process of handling non-English inputs in an LLM can be roughly divided into three stages from the bottom layers to the top layers: Language Conversion, English Thinking and Language Reversion.   
    }
  \label{fig:fig2}
\end{figure*}

\section{Method}

In this section, we describe the proposed deep supervision-based fine-tuning approach. Figure \ref{fig:fig2} illustrates the overall process of our method. The process of handling non-English inputs in an LLM can be roughly divided into three stages from the bottom layers to the top layers: (1) Language Conversion: The model interprets the non-English query and converts the multilingual input into English (or an English representation space); (2) English Thinking: The model employs English for thinking and solving the task; (3) Language Reversion: The model converts the reasoning results back into the target language, consistent with the input. We propose DFT to guide the model's internal information transformation. Specifically, we introduce additional supervision at the top and middle layers of the model to enhance the model's ability to convert non-English inputs into English and to improve its logical reasoning in English. We propose two types of supervision: logits-based  and feature-based.

\subsection{Language Conversion Constraints}

When processing non-English inputs, the initial layers of large language models are primarily tasked with converting the input query into an internal English representation that the model can effectively process. Due to the significant differences in grammar, vocabulary, and linguistic structures across languages, this conversion process is inherently challenging. Failure to accurately perform this conversion can lead to semantic biases or errors in subsequent reasoning and generation stages. Thus, it is critical that the model excels in language conversion within its early layers.

To enhance the model’s language conversion capabilities, we propose two constraint methods: logits supervision and feature supervision.

\textbf{Logits-Based:} For transformer-based LLMs, each layer outputs a hidden representation of shape $(batch\_size \times sequence\_length \times hidden\_size)$. The typical training strategy feeds the hidden representation from the final layer into a linear matrix $W^{out}$ of shape $(hidden\_size \times vocab\_size)$, projecting the final hidden representation into the vocabulary space and predicting the probability distribution over the vocabulary using softmax. This distribution is then compared with the ground truth to calculate the loss. However, since the hidden states have the same shape across all layers, we can apply the $W^{out}$ and softmax operations at any layer to make predictions. To enforce language conversion, we apply $W^{out}$ at an internal layer $i$ in the model (as shown in Figure 2b) and compute the probability distribution. Then calculate the loss with a parallel English query, thereby forcing the model to convert the target language query into its English version in the early layers. Where the logits-based language conversion loss $\mathcal{L}_{LC}$ is computed as:
\begin{equation}
    \label{eq:loss1_logits}
    \mathcal{L}_{LC} =  - \log P(x^{en}|x^{tgt}; \theta_{1:i},W^{out}),
\end{equation}
$\theta_{1:i}$ represents the parameters of the first $i$ layers of the model. We use the LLM's head as $ W^{out} $ and restrict $ W^{out} $ to only participate in the forward pass during the language conversion stage, without being updated.

\textbf{Feature-Based:} Another approach to enhancing the language conversion process in the first stage is through feature alignment. During instruction tuning in the target language, semantically equivalent English querys are also fed into the model. The hidden representations of these inputs are then extracted, and a similarity loss is calculated to ensure alignment between the English and non-English representations. The feature-based language conversion loss function $\mathcal{L}_{LC}$ can be defined as:
 
\begin{equation}
\mathcal{L}_{LC} = 1-  \left( \frac{h^{en}_i \cdot h^{tgt}_i}{\|h^{en}_i\| \|h^{tgt}_i\|} \right),
\end{equation}
where $i$ represents the critical layer in the first stage, $ h^{en} $ and $ h^{tgt} $ represent the hidden states of the same semantic input in English and the target language, respectively. Compared to the logits-based supervision strategy, feature supervision is relatively more relaxed. It does not require the target language input to be strictly converted into English but instead aligns the representations across languages.

By applying constraints at this stage, non-English inputs can be more accurately transformed into English, thereby laying a strong foundation for subsequent reasoning and output generation. 

\subsection{English Thinking Constraints}

After language conversion, the model needs to further process the input information and reason how to provide an appropriate response. 
Therefore, we add supervisory signals to constrain the model's reasoning ability in English. Similar to the previous section, we implement with both logits constraint and feature constraint methods. 

\textbf{Logits-Based:} For the logits constraint, we output probabilities from the hidden vectors at the end of the second stage and compute the loss, aiming for the model to provide corresponding English answers based on the input target question. The logits-based english thinking loss function $\mathcal{L}_{ET}$ is formulated as follows:

\begin{equation}
    \label{eq:loss1_logits}
    \mathcal{L}_{ET} =  - \log P(y^{en}|x^{tgt}; \theta_{1:j},W^{out}),
\end{equation}
where $j$ represents the critical layer in the second stage responsible for reasoning in English. This constraint forces the model to focus on producing the correct English reasoning path, ultimately leading to accurate final outputs.

\textbf{Feature-Based:} For the feature constraint, as shown in the Figure \ref{fig:fig2}, this is implemented by feeding semantically equivalent English and non-English questions into the model. For the English input, we extract the features from the top layers of the model, while for the non-English input, we extract the features from the critical layer of the English Thinking stage, where the reasoning in English is assumed to occur. The alignment of these features is crucial for ensuring that the model’s internal representations remain consistent across language.

\begin{equation}
\mathcal{L}_{ET} = 1-  \left( \frac{h^{en}_L \cdot h^{tgt}_j}{\|h^{en}_L\| \|h^{tgt}_j\|} \right),
\end{equation}
where $L$ represents the final layer, $j$ represents the critical layer in the second stage. This constraint helps the model to align its reasoning process in English with the semantic content derived from the target language input, ensuring that the reasoning remains accurate and consistent in the model.

\subsection{Training Objective}

The total loss function is defined as:
\begin{equation}
    \label{eq:loss1_logits}
    \mathcal{L} = \mathcal{L}_{TFT}+\mathcal{L}_{LC} + \mathcal{L}_{ET},
\end{equation}
where 
$\mathcal{L}_{LC}$ is either logits-based or feature-based, and $\mathcal{L}_{ET}$
is also either logits-based or feature-based.

\begin{table*}[ht]
\renewcommand\tabcolsep{4.0pt}
\centering
\begin{tabular}{cccccccccccccc}
\toprule
\multirow{2}{*}{\textbf{}} & \multirow{2}{*}{\textbf{Method}} & \multicolumn{3}{c}{\textbf{xquad}} & \multicolumn{3}{c}{\textbf{mlqa}} & \multicolumn{3}{c}{\textbf{truthfulQA}} & \multicolumn{3}{c}{\textbf{mkqa}} \\
 &  & \textbf{vi} & \textbf{ar} & \textbf{zh} & \textbf{vi} & \textbf{ar} & \textbf{zh} & \textbf{vi} & \textbf{ar} & \textbf{zh} & \textbf{vi} & \textbf{ar} & \textbf{zh} \\
\midrule
\midrule
\multicolumn{14}{c}{\textit{Performance on LLaMA-2-7b}} \\
\multirow{3}{*}{Baselines}   & SFT & 23.11 & 19.42 & 3.34 & 22.79 & 19.18 & 6.91 & \textbf{29.37} & \textbf{29.11} & \textbf{30.49} & 33.16 & 32.84 & 33.51  \\ & TFT & 27.36 & 21.23 & 23.41 & 28.80 & 21.62 & 23.86 & 28.54 & 27.04 & 26.40 & 38.59 & 35.02 & 40.70  \\ 

 & SDRRL & 28.79 & 25.75 & 22.01 & 29.59 & 27.26 & 24.67 & 28.03 & 28.11 & 28.30 & 39.20 & 35.81 & 40.63 \\
 \midrule
\multirow{3}{*}{Ours} & DFT-logits & \textbf{29.65} & \textbf{27.57} & \textbf{24.77} & 31.54 & 28.26 & \textbf{25.68} & 28.64 & 28.81 & 28.93 & \textbf{39.64} & \textbf{35.84} & 40.87 \\
 & DFT-feature & 28.27 & 26.35 & 24.34 & \textbf{32.19} & \textbf{29.51} & 25.29 & 28.61 & 28.34 & 28.00 & 39.36 & 35.65 & \textbf{40.90} \\
\midrule
\multicolumn{14}{c}{\textit{Performance on Gemma-2-2b}} \\
\multirow{3}{*}{Baselines} & SFT & 19.02 & 18.82 & 9.74 & 20.21 & 20.45 & 11.79 & \textbf{30.45} & \textbf{28.98} & \textbf{27.79} & 36.04 & 33.69 & 32.46 \\
& TFT & 17.58 & 24.73 & 15.95 & 20.26 & 25.31 & 18.26 & 26.44 & 26.26 & 24.87 & 36.84 & 34.49 & 40.20 \\
& SDRRL & 17.55 & 18.16 & 12.03 & 19.89 & 23.91 & 14.24 & 26.62 & 26.52 & 26.39 & 37.69 & 34.51 & 40.75 \\
\midrule
\multirow{2}{*}{Ours} & DFT-logits & 21.23 & 25.79 & \textbf{17.37} & \textbf{24.36} & 27.51 & \textbf{20.13} & 27.34 & 26.78 & 27.16 & \textbf{38.59} & \textbf{34.81} & 40.84 \\
& DFT-feature & \textbf{22.78} & \textbf{27.38} & 16.94 & 22.19 & \textbf{30.18} & 19.51 & 27.88 & 27.64 & 26.90 & 38.04 & 34.73 & \textbf{40.87} \\

\bottomrule

\end{tabular}
\caption{Results of baselines and our method on multilingual question and answer benchmark. Bold indicates the best result of all methods. Our method outperforms the baselines in almost all languages.}
\label{tab:QA}
\end{table*}

\begin{table*}[ht]
\renewcommand\tabcolsep{4.0pt}
\centering
\begin{tabular}{cccccccccccc}
\toprule

\multirow{2}{*}{\textbf{}} & \multirow{2}{*}{\textbf{Method}} & \multicolumn{3}{c}{\textbf{xnli}} & \multicolumn{2}{c}{\textbf{xcopa}} & \multicolumn{2}{c}{\textbf{xstory\_cloze}} & \multicolumn{3}{c}{\textbf{mmlu}} \\
 &  & \textbf{vi} & \textbf{ar} & \textbf{zh} & \textbf{vi} &  \textbf{zh}  & \textbf{ar} & \textbf{zh} & \textbf{vi} & \textbf{ar} & \textbf{zh} \\
\midrule
\midrule
\multicolumn{12}{c}{\textit{Performance on LLaMA-2-7b}} \\
\multirow{3}{*}{Baselines}   & SFT & 36.59 & \textbf{35.38} & 36.39 & 63.00 & 65.00 & 49.70 & 59.56 & 29.37  & 27.6 & 30.49 \\

& TFT & \textbf{44.14} & 33.73 & 36.87 & 66.00 & 65.00 & 57.51 & 64.06 & \textbf{33.23}  & 28.05 & 32.34 \\

 & SDRRL & 43.41 & 33.98 & 37.43 & 65.40 & 65.20 & 58.17  & 64.53 & 31.91 & 27.47 & 32.74 \\
 \midrule
\multirow{3}{*}{Ours} & DFT-logits & 43.37 & 34.06  & 36.94 & 66.20 & 66.40 & \textbf{58.78} & \textbf{64.93} & 32.60 & \textbf{29.06} & \textbf{33.19} \\
 & DFT-feature & 43.50 & 34.38 & \textbf{37.64} & \textbf{66.60} & \textbf{68.60}  & \textbf{58.78} & 64.40 & 32.19 & 28.86 & 32.82 \\
\midrule
\multicolumn{12}{c}{\textit{Performance on Gemma-2-2b}} \\
\multirow{3}{*}{Baselines}  &SFT & 41.33 & 35.62 & \textbf{40.36} & \textbf{68.60} & 70.40 & 61.02 & \textbf{68.23}  & 36.07 & 31.16 & 36.22 \\
& TFT & 43.05 & 33.94 & 36.63 & 67.00 & 70.40 & 61.48 & 64.88  & \textbf{36.95} & \textbf{32.71} & \textbf{36.37} \\
& SDRRL & 42.25 & 36.31 & 39.20 & 65.40 & 66.60 & 61.09 & 63.20 & 34.47 & 31.31 & 34.23 \\
\midrule
\multirow{2}{*}{Ours} & DFT-logits & \textbf{43.29} & 35.14 & 38.46  & 68.40 & \textbf{70.60} & \textbf{62.08} & 66.91 & 36.89 & 31.84 & 35.30 \\
& DFT-feature & 43.25 & \textbf{36.73} & 39.74 & 67.20 & 70.20 & 61.88  & 66.25 & 36.70 & 31.63 & 34.43 \\
\bottomrule

\end{tabular}
\caption{Results of baselines and our method on multilingual understanding  benchmark. Our method outperforms the baselines in almost all languages. }
\label{tab:NLU}
\end{table*}

\section{Experiment}
\subsection{Setup}
We use LLaMA-2-7B \cite{touvron2023llama} and Gemma-2-2B \cite{gemma_2024} as the base models. The training data consists of Stanford Alpaca instruction data \cite{taori2023stanford} and its translations in the target languages, which include Chinese (zh), Vietnamese (vi), and Arabic (ar). For the translated data, we directly used publicly available datasets from \cite{zhu2023extrapolating}. Our code implementation is based on stanford\_alpaca \footnote{\url{https://github.com/tatsu-lab/stanford_alpaca}.}.
All experiments were conducted on 8 $\times$ A100 GPUs with a batch size of 128. The models were trained for 3 epochs with a learning rate of 2e-5. To accelerate training, we utilized the FSDP training strategy \cite{zhao2023pytorch}.
\subsection{Comparison of Methods}
\begin{itemize}
    \item \textbf{SFT} \cite{ouyang2022training}, which is instruction-tuned with English instruction datasets.
    \item \textbf{TFT} \cite{zhu2023extrapolating}, which is instruction-tuned using the original English instruction datasets translated into the target languages.
    \item \textbf{SDRRL} \cite{zhang2024enhancing}, which is a method based on Self-Distillation. Besides using English instruction-tuning data and its multilingual code-switching extensions, it also incorporates partially translated data and completion data for fine-tuning. 
    \item \textbf{DFT-logits}, our method that applies logits-based supervision to guide the model's intermediate layers.
    \item \textbf{DFT-feature}, our method that uses feature alignment to maintain consistent internal representations between English and tgt language. 
\end{itemize}

\subsection{Evaluation Dataset}

\begin{itemize}
\item\textbf{XQUAD (Cross-lingual Question Answering Dataset):}
XQUAD \cite{artetxe2019cross} is a high-quality cross-lingual question answering dataset containing 240 paragraphs and 1,190 question-answer pairs, which have been manually translated into 10 languages. 

\item\textbf{MLQA (Multilingual Question Answering):}
MLQA \cite{lewis2019mlqa} is a multilingual question answering dataset covering 7 languages. Each question in the dataset is accompanied by a paragraph and an answer in the corresponding language.

\item\textbf{MKQA (Multilingual Knowledge Questions and Answers):}
The MKQA \cite{longpre2021mkqa} dataset contains 2,600 common sense question-answer pairs across 26 languages. 

\item\textbf{TruthfulQA}:
TruthfulQA \cite{lin2021truthfulqa} includes questions from various domains, specifically designed to test the truthfulness and accuracy of models when answering complex questions.

\item\textbf{XNLI (Cross-lingual Natural Language Inference):}
XNLI \cite{conneau2018xnli} is a widely used language understanding dataset to evaluate models' performance in cross-lingual inference tasks.

\item\textbf{XCOPA (Cross-lingual Choice of Plausible Alternatives):}
XCOPA \cite{ponti2020xcopa} is a benchmark designed to evaluate the ability of models to apply commonsense reasoning, requiring both world knowledge and the ability to generalize it to new languages.

\item\textbf{XStoryCloze (Cross-lingual Story Cloze Test)}
XStoryCloze \cite{lin2022few} is a cross-lingual dataset for evaluating models' ability to understand stories and generate plausible endings.

\item\textbf{MMLU (Massive Multitask Language Understanding)}
MMLU \cite{hendrycks2020measuring} is a large-scale multitask language understanding dataset covering multiple domains (such as history, geography, science, law, etc.) and various languages.

\end{itemize}

For all evaluation datasets, we conducted tests using a zero-shot setting. We used the F1 score for XQuAD, MLQA, and MKQA, and the MC1 metric for TruthfulQA. For other NLU datasets, accuracy was used as the evaluation metric.

\subsection{Main Results}

Table \ref{tab:QA} and Table \ref{tab:NLU} present the results on multilingual QA and NLU tasks, respectively. From the experimental results, we can observe that:
(1) Our method outperforms the baselines in almost all languages for both QA and NLU tasks. This indicates that our approach successfully enhances the model's capabilities in the target language by guiding the internal workflow. (2) The improvement is more pronounced in QA tasks, as our method is specifically designed for tasks where both the input and output are in the target language, making it better suited for QA scenarios in the target language. (3) Fine-tuning the model solely on English instruction data (SFT) outperformed all other results fine-tuned on target language instruction datasets for TruthfulQA. This suggests that for the MC1 metric in TruthfulQA, generation capabilities in the target language are less important. (4) We observed that DFT (feature) performs better than DFT (logits) on understanding tasks, possibly because the stricter logits-based supervision is more suitable for generation tasks, whereas the feature-based supervision offers better generalization across different types of tasks.

\subsection{Ablation Study}

We further analyzed the effects of applying supervision over either the first or the second stage. We compared the performance of separately adding Language Conversion supervision (LC) and English Thinking supervision (ET) on 8 Chinese datasets. The results, based on LLaMA2, are shown in the Table \ref{tab:ablation}, where the supervision types are represented as "logits-based / feature-based."

From the experimental results, we can observe that: (1) Adding supervision during the English Thinking stage yielded more significant improvements, indicating that aligning responses has a greater impact. (2) Logits-based supervision led to greater improvements in QA tasks, while feature-based supervision was more beneficial for understanding tasks. This may be because generation tasks require stricter supervision signals. (3) Logits-based language conversion supervision caused performance drops in some tasks, suggesting that strong supervision over the earlier layers may harm the model's original capabilities.

\begin{table}[!t]
\renewcommand\tabcolsep{4.0pt}
\centering
\begin{tabular}{cccccc}
\toprule
& TFT & \multicolumn{1}{c}{+LC} & \multicolumn{1}{c}{+ET} \\
\midrule
\textbf{XQUAD} & 23.41 & \textbf{24.75} / 23.99 & 24.47 / 24.26   \\
\textbf{MLQA} & 23.86 & 25.62 / 24.27  & \textbf{25.68} / 24.80  \\
\textbf{TruthfulQA} & 26.40 & 27.11 / 27.51 & \textbf{28.93} / 28.00   \\
\textbf{MKQA} & 40.70 & 40.73 / 40.81 & \textbf{40.83} / 40.70  \\
\textbf{XNLI} & 36.87 & 36.27 / 36.59 & 37.61 / \textbf{37.75}  \\
\textbf{XCOPA} & 65.00 & 65.60 / 66.80  & 66.40 / \textbf{68.40}  \\
\textbf{Xstory\_Cloze} & 64.06 & 63.60 / 64.92 & \textbf{64.95} / 64.26   \\
\textbf{MMLU} & 32.34 & 32.75 / \textbf{33.73}  & 33.29 / 32.85  \\
\bottomrule
\end{tabular}
\caption{Results on 8 Chinese evaluation datasets with separately added language conversion supervision (LC) and english thinking supervision (ET). The supervision types are represented in ``logits-based / feature-based" form.}
\label{tab:ablation}
\end{table}

\begin{figure}[!t]
  \centering
    \includegraphics[clip,width=0.9\columnwidth,]{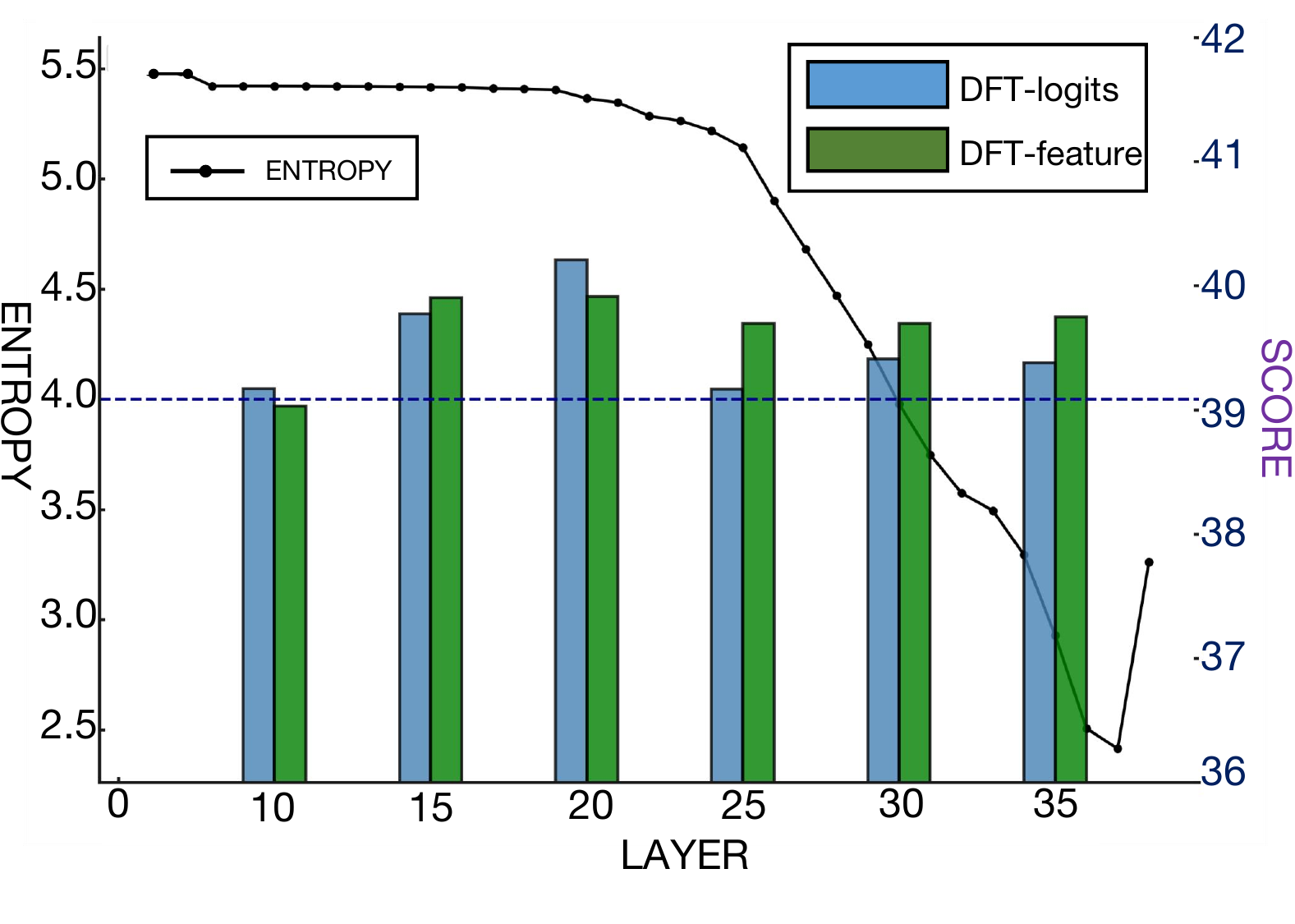}
    \caption{The bars in the figure represent the results of DFT-logits and DFT-feature on all evaluation datasets, with English Thinking supervision applied at different layers. The target language is Chinese, and the base model used is LLaMA-2-7b. The dashed line indicates the results of the TFT method. The average scores across various datasets are reported. The broken line represents the change in entropy as the layer depth increases.}
  \label{fig:entropy}
  
\end{figure}

\subsection{Entropy-Based Critical Layer Selection}
To implement our approach, it is crucial to accurately identify the critical layers that separate the different stages, particularly when determining the 
$i-th$ and $j-th$ layers as shown in the Figure \ref{fig:fig2}.
Although previous works \cite{wendler2024llamas, zhao2024large} have discovered the mechanisms of information transformation within models when processing non-English inputs, identifying the critical layers at different stages remains challenging.

Entropy, in the context of information theory, is a measure of uncertainty or randomness in the information being processed. In neural networks, entropy can help identify where significant transformations or reductions in uncertainty occur. We observed that when LLMs processes non-English inputs, there are two significant drops in entropy (as shown in Figure \ref{fig:fig2} (a)) . 
These drops indicate key points where the model undergoes substantial information transformation. Based on these observations, we hypothesize that the initial entropy drop  corresponds to the model processing the non-English input into an English representation that it can handle. After this, the model transitions to the English reasoning and processing stage. The subsequent entropy drop marks the model gradually completing the reasoning process and progressively forming the final output.

To validate the effectiveness of our hypothesis, we applied English Thinking supervision at layers 5, 10, 15, 20, 25,  30 of LLaMA-2-7b and fine-tuned the model on the Chinese instruction dataset. Figure \ref{fig:entropy} shows the average results across 8 Chinese datasets using the DFT-logits and DFT-feature methods. We observed that: (1) Entropy first drops at layer 2, stabilizes for a period, and then begins to drop again around layer 15. (2) Both DFT-logits and DFT-feature achieve better performance when applied at layer 15, indicating that our hypothesis is valid. (3) For the DFT-logits method, performance declines when supervision is applied at later layers, possibly because the model has already started the language reversion process (converting English back to the target language). Adding constraints to predict English results at this stage may interfere with the model's generation of the target language. In contrast, the relatively more relaxed DFT-feature method performs better at later layers.

Although this hypothesis is rough (as different tokens may exhibit different behaviors), it provides us with guidance for selecting the critical layer at each stage.

\subsection{Analysis of Representation Alignment}

We used the t-SNE \cite{van2008visualizing} method to visualize the representations of input sentences to analyze the impact of DFT on aligning cross-lingual representations. 

Specifically, we encoded parallel English and Chinese  sentences from the FLORES-200 dataset and obtain sentence representations by the mean pooling method using the representation for each token.

\begin{figure}[!t]
  \centering
    \includegraphics[clip,width=1.0\columnwidth,]{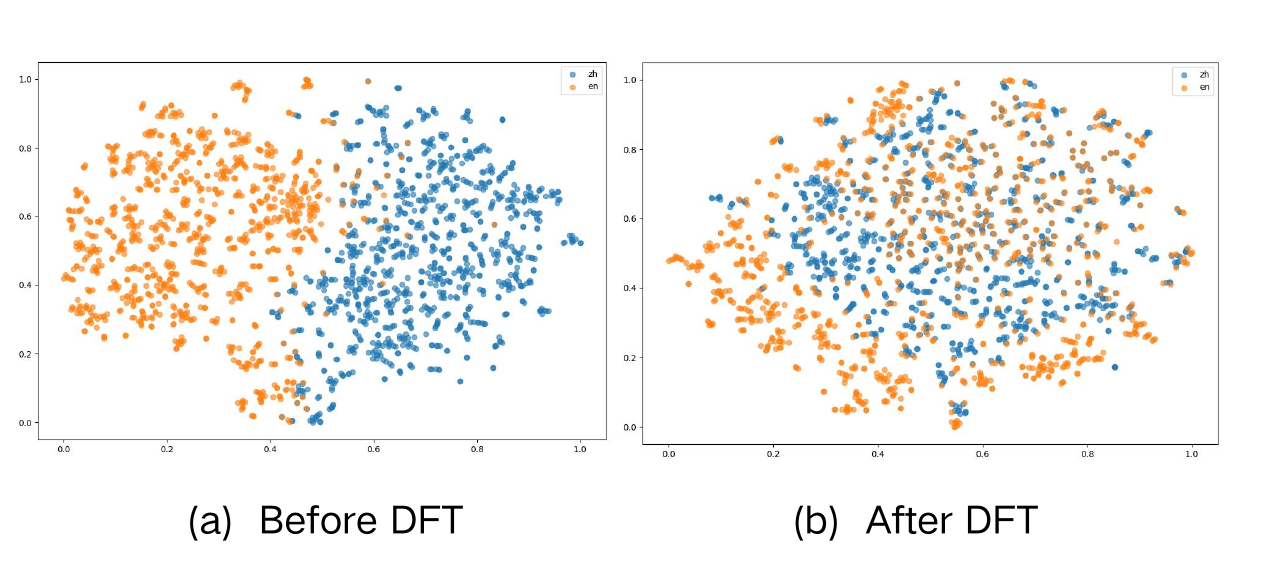}
    \caption{t-SNE visualizations of sentence representations from FLORES-200 dataset by LLaMA-2 before and after applying DFT.
    }
  \label{fig:tsne}
\end{figure}

The results are shown in Figure \ref{fig:tsne}. In the vanilla model, the representations of the two languages are far apart. After applying the DFT method, they become more aligned. This indicates that our method can help bring the target language representations closer to the English representations.

\subsection{Analysis of Translation Task}

The model's performance on translation tasks can reflect its overall language conversion and generation capabilities. Therefore, although cross-lingual generation tasks do not align with the workflow of our method, we still analyzed our approach on translation tasks. We compared the performance of separately adding Language Conversion supervision and English Thinking supervision,  and the results are shown in Table 3.

From the experimental results, we can observe that: (1) Adding logits-based supervision led to a catastrophic drop in the en-zh direction. This suggests that strict supervision of target language conversion into English within the model's internal layers leads to a decline in the model's overall ability to convert English to the target language. (2) Adding feature-based supervision significantly improved translation results for both en-zh and zh-en directions, indicating that aligning representations between languages is beneficial for cross-lingual tasks.

In fact, our method is better suited for scenarios where both the input and output are in the target language, rather than for cross-lingual tasks. Nevertheless, DFT-feature still achieved strong performance on translation tasks, indicating that our approach has broad potential for application.

\begin{table}[ht]
\renewcommand\tabcolsep{4.0pt}
\centering
\begin{tabular}{cccccc}
\toprule
& TFT & \multicolumn{1}{c}{+LC} & \multicolumn{1}{c}{+ET} \\
\midrule
\textbf{en-zh} & 68.37 & 36.79 / \textbf{75.62} & 46.50 / 74.49   \\
\textbf{zh-en} & 63.12 & 64.10 / 82.58 & 63.68 / \textbf{83.98}  \\
\bottomrule
\end{tabular}
\caption{Translation performance on FLORES-200 with separately added language conversion supervision (LC) and english thinking supervision (ET), evaluated using COMET score. 
}
\label{tab:translation}
\end{table}
\section{Related Work}

\subsubsection{Aligning Non-English Capabilities of Large Language Models}

To enhance the non-English capabilities of LLMs, researchers have explored several approaches. Pre-training LLMs on diverse multilingual datasets has proven effective in improving multilingual performance. However, this approach requires the collection of large amounts of data and significant computational resources \cite{le2023bloom,cui2023efficient,Chinese-Mixtral-8x7B}. Instruction fine-tuning on translation datasets has also been successful in enhancing non-English performance\cite{li2023bactrian,zhu2023extrapolating,zhang2024enhancing,xu2023language,li2024x}.
In the inference stage, cross-lingual transfer methods, such as leveraging knowledge from resource-rich languages and using self-translation prompts, have been effective \cite{qin2023cross,xu2023language,huang2023not}. Additionally, \citet{li2024improving} aligned English representations to enable the model to fully leverage its English capabilities when processing non-English inputs.

\subsubsection{Deep Supervision Networks}

\citet{lee2015deeply} first proposed the deeply supervised network, where auxiliary classifiers are added on various intermediate layers, and each classifier contributes to the overall loss during training. 
\citet{huang2022non} enhanced machine translation performance by predicting outputs layer by layer in a non-autoregressive manner. \citet{elbayad2019depth} improved decoding efficiency by making predictions at different layers based on token prediction differences. 

Our method draws on the implementation of deeply supervised networks by introducing supervision over the middle part of the model. Our goal is to guide the internal information transformation process within the model.

\section{Conclusion and Future Work}
In this work, we propose Deep supervision Fine-Tuning, effectively enhancing the multilingual capabilities of English-dominated LLMs. 
Our method guides the workflow of LLMs when processing non-English inputs by adding cross-lingual supervision over intermediate layers, constraining models to achieve more accurate language conversion and obtain more precise intermediate results. Experimental results demonstrate that our method significantly improves performance on
various multilingual tasks. 

We devise an entropy-based method for critical layer selection and have preliminarily validated its effectiveness. However, variations among tokens within samples suggest that this guidance is still imprecise. We will further explore this issue in the future.

\section{Acknowledgments}
Bing Qin and Xiaocheng Feng are the co-corresponding authors of this work. We thank the anonymous reviewers for their insightful comments. This work was supported by the National Key R\&D Program of China via grant No. 2021ZD0112905, National Natural Science Foundation of China (NSFC) via grant 62276078, the Key R\&D Program of Heilongjiang via grant 2022ZX01A32, the Major Key Project of PCL via grant No. PCL2023A09, the International Cooperation Project of PCL, PCL2022D01 and the Fundamental Research Funds for the Central Universities (Grant No.HIT.OCEF.2023018).

\bibliography{aaai25}

\end{document}